\documentclass{article} 
\usepackage{iclr2021_conference,times}

\usepackage{amsmath,amsfonts,bm}

\def\eqref#1{equation~\ref{#1}}

\def\1{\bm{1}}

\DeclareMathAlphabet{\mathsfit}{\encodingdefault}{\sfdefault}{m}{sl}
\SetMathAlphabet{\mathsfit}{bold}{\encodingdefault}{\sfdefault}{bx}{n}

\usepackage{hyperref}
\usepackage{url}
\usepackage{graphicx}
\iclrfinalcopy

\title{Adversarial autoencoders and adversarial LSTM for improved forecasts of urban air pollution simulations}

\author{%
César Quilodrán-Casas\thanks{Corresponding author}\thanks{Code available from \url{https://github.com/c-quilo/adversarial-AE-LSTM/}}\\
  Data Science Institute\\
  Imperial College London\\
  \texttt{caq13@imperial.ac.uk} \\
  \And
 Rossella Arcucci\\
    Data Science Institute\\
    Leonardo Centre, Imperial Business School\\
    Imperial College London\\
  \texttt{r.arcucci@imperial.ac.uk}\\
    \AND
 Laetitia Mottet \\
Department of Earth Science \& Engineering\\
  Imperial College London\\
  \texttt{l.mottet@imperial.ac.uk}\\
    \And
Yike Guo \\
Data Science Institute\\
Imperial College London\\
\texttt{y.guo@imperial.ac.uk}\\
  \AND
  Christopher C. Pain\\
Department of Earth Science \& Engineering\\
  Imperial College London\\
  \texttt{c.pain@imperial.ac.uk}\\

}

\begin{document}

\maketitle

\begin{abstract}

This paper presents an approach to improve the forecast of computational fluid dynamics (CFD) simulations of urban air pollution using deep learning, and most specifically adversarial training. This adversarial approach aims to reduce the divergence of the forecasts from the underlying physical model. Our two-step method integrates a Principal Components Analysis (PCA) based adversarial autoencoder (PC-AAE) with adversarial Long short-term memory (LSTM) networks. Once the reduced-order model (ROM) of the CFD solution is obtained via PCA, an adversarial autoencoder is used on the principal components time series. Subsequentially, a Long Short-Term Memory network (LSTM) is adversarially trained on the latent space produced by the PC-AAE to make forecasts. Once trained, the adversarially trained LSTM outperforms a LSTM trained in a classical way. The study area is in South London, including three-dimensional velocity vectors in a busy traffic junction.

\end{abstract}

\section{Introduction}

Data-driven approaches can be seen as attractive solutions to produce reduced-order models (ROMs) of Computational Fluid Dynamics (CFD) simulations. Moreover, forecasts produced by ROMs are obtained at a fraction of the cost of the original CFD model solution when used together with a ROM. Recurrent neural networks (RNN) have been used to model and predict temporal dependencies between inputs and outputs of ROMs. Non-intrusive ROMs and RNNs have been used together in previous studies, e.g. \citep{quilodran2018fast, quilodran2021adversarially, reddy2019reduced} where the surrogate forecast systems can easily reproduce a time-step in the future accurately. However, when the predicted output is used as an input for the prediction of the subsequent time sequence, the results can detach quickly from the underlying physical model solution when encountering out-of-distribution data.

Our framework relies on adversarial training to improve the longevity of the surrogate forecasts. \citet{goodfellow2014generative} introduced the idea of adversarial training and adversarial losses which can also be applied to supervised scenarios and have advanced the state-of-the-art in many fields over the past years \citep{dong2019towards, wang2019improving}. Additionally, robustness
may be achieved by detecting and rejecting adversarial examples by using adversarial training \citep{shafahi2019adversarial, meng2017magnet}. Data-driven modelling of nonlinear fluid flows incorporating adversarial networks have been successfully being studied previously \citep{cheng2020data, xie2018tempogan}.

This extended abstract applies PC-based adversarial autoencoder \citep{makhzani2015adversarial} and adversarial training to a LSTM network, based on a ROM \citep{casas2020urban, phillips2020autoencoder} of an urban air pollution simulation in an unstructured mesh. The motivation of using adversarial autoencoders relies in the ability of the adversarial training to match the aggregated posterior of the encoder with an arbitrary prior distribution. This process aids the LSTM to be trained in the matched latent space to avoid producing out-of-distribution forecast samples. The robustness added by the adversarial training allows us to reduce the divergence of the forecast prediction over time and better compression from full-space to latent space.
\section{Methods}
\begin{figure*}[ht]
\centering
\includegraphics[width=0.9\linewidth]{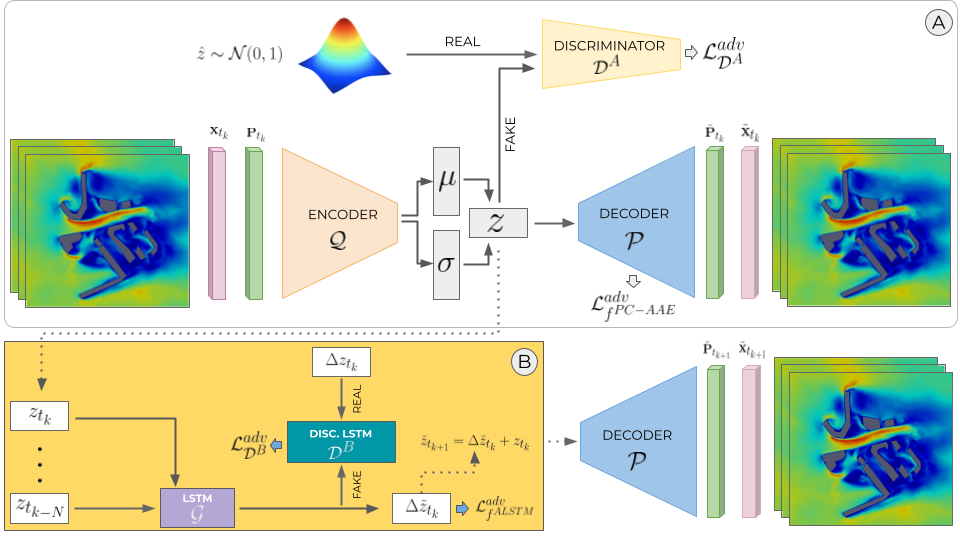}
\caption{Workflow. Network A: PC-based adversarial autoencoder. Network B: Adversarial LSTM. Solid lines input within the same network, dashed lines input to another network.}
\label{fig:workflow}
\end{figure*}
\subsection{Principal Components Analysis}

As described by \citet{lever2017points}, PCA is an unsupervised learning method that simplifies high-dimensional data by transforming it into fewer dimensions. Let $\mathbf{x}=\left\{ \mathbf{x}_t \right\}_{t=1,\dots,n}$ with $\mathbf{x}\in \Re^{n \times m}$, with $n<m$, denotes the matrix of the model vectors at each time step. The PCA consists in decomposing this dataset as $\mathbf{x} = \mathbf{P} \mathbf{\Pi} + \mathbf{\bar{x}}$ where $\bf{P} \in \it{\mathbb{\Re}^{n\times n}}$ are the principal components of $\mathbf{x}$; $\bf{\Pi} \in \it{\mathbb{\Re}^{n\times m}}$ are the Empirical Orthogonal Functions; and $\mathbf{\bar{x}}$ is the mean vector of the model. The dimension reduction of the system comes from truncating $\mathbf{P}$ at the first $\tau$ PCs  as $\mathbf{x}_{\tau} = \mathbf{P}_{\tau} \mathbf{\Pi}_{\tau} + \mathbf{\bar{x}}$, with $\mathbf{P}_{\tau} \in \Re^{n\times \tau}$ and $\mathbf{\Pi}_{\tau} \in \Re^{\tau \times m}$.

\subsection{PC-based adversarial autoencoder}
\label{sec:aae}
Once the PCs are obtained, they are utilised to train an adversarial autoencoder (AAE). The functional of our PC-based adversarial autoencoder (PC-AAE) is defined as:

\begin{equation}
    f^{PC-AAE}:\mathbf{P}_{t_k} \to \tilde{\mathbf{P}}_{t_k}
\end{equation}
where $\mathbf{P_{t_k}}$ are the scaled principal components time series between -1 and 1 at time-level $k$. The autoencoder consists of an encoder $\mathcal{Q}$ and a decoder $\mathcal{P}$, both mirrored fully-connected networks, where $\tilde{\mathbf{P}}_{t_k} = \mathcal{P}(\mathcal{Q}(\mathbf{P_{t_k}}))$.

Let $q(\mathbf{z}|\mathbf{P})$ and $p(\tilde{\mathbf{P}}|\mathbf{z})$ be the encoding and decoding distributions, respectively. As suggested by \citet{makhzani2015adversarial}, we use a Gaussian posterior and assume that $q(\mathbf{z}|\mathbf{P})$ is a Gaussian distribution, where its mean and variance are predicted by the encoder $\mathcal{P}$: $\mathbf{z}\sim\mathcal{N}(\mu(\mathbf{P}), \sigma(\mathbf{P}))$. This is achieved by adding two dense layers of means $\mu$ and $log\ \sigma$, (see Figure \ref{fig:workflow}) to the final layer of the encoder $\mathcal{Q}$, and return $\mathbf{z}$ as a vector of samples. To ensure that $\mathbf{z}\sim q(\mathbf{z}) = \mathcal{N}(\mu,\sigma^{2})$, the aggregated posterior, we use the reparameterisation trick described by \citet{kingma2013auto} for backpropagation through the network $\mathbf{z}=\mu+\sigma\odot\mathbf{\epsilon}$, where $\mathbf{\epsilon}$ is an auxiliary noise variable $\mathbf{\epsilon} \sim \mathcal{N}(0, \mathbf{I})$.

The adversarial training of PC-AAE includes a discriminator $\mathcal{D}^{A}$ to distinguish between the real samples, given by an arbitrary prior $p(\mathbf{z}) \sim \mathcal{N}(0,\mathbf{I})$ and $q(\mathbf{z})$. Therefore, the adversarial autoencoder is regularised by matching $p(\mathbf{z})$ to $q(\mathbf{z})$. The $\mathbf{P}$ are fed to the discriminator as real sequences (ground truth). Let, $\mathcal{D}^{A}(\alpha, \beta)$ represent the discriminator function with an input $\alpha$ and a target label $\beta$ such that, for $\alpha=\hat{\mathbf{z}} \sim p(\mathbf{z})$, $\beta=1$ and for $\alpha=\mathbf{z} \sim q(\mathbf{z})$, $\beta=0$ . The training of $\mathcal{D}^{A}$ is based on the minimisation of the binary cross-entropy loss ($\mathcal{L}^{bce}$), using the Nesterov Adam optimizer (Nadam) \citep{dozat2016incorporating}. The adversarial losses $\mathcal{L}^{adv}$ for $\mathcal{D}^{A}$ and $f^{PC-AAE}$ are then defined as:

\begin{align}
    \mathcal{L}^{adv}_{\mathcal{D}^{A}}(\mathbf{P}) &= \mathcal{L}^{bce}_{\hat{\mathbf{z}}\sim p(\mathbf{z})}(\mathcal{D}^{A}(\hat{\mathbf{z}}), 1)) + \mathcal{L}^{bce}_{\mathbf{z}\sim q(\mathbf{z})}(\mathcal{D}^{A}(\mathbf{z}), 0))\\
    \mathcal{L}^{adv}_{f^{PC-AAE}}(\mathbf{P}) &= \mathcal{L}^{bce}_{\mathbf{z}\sim q(\mathbf{z})}(\mathcal{D}^{A}(\mathbf{z}), 1)) + \mathcal{L}^{mse}(\tilde{\mathbf{P}}, \mathbf{P})
\end{align}
where $\mathcal{L}^{mse}$ is the mean squared error (mse) between $\tilde{\mathbf{P}}$ and $\mathbf{P}$.

\subsection{Adversarially long short-term memory network}
\label{sec:advLSTM}
\begin{figure*}[t]
\centering
\includegraphics[width =0.9\textwidth]{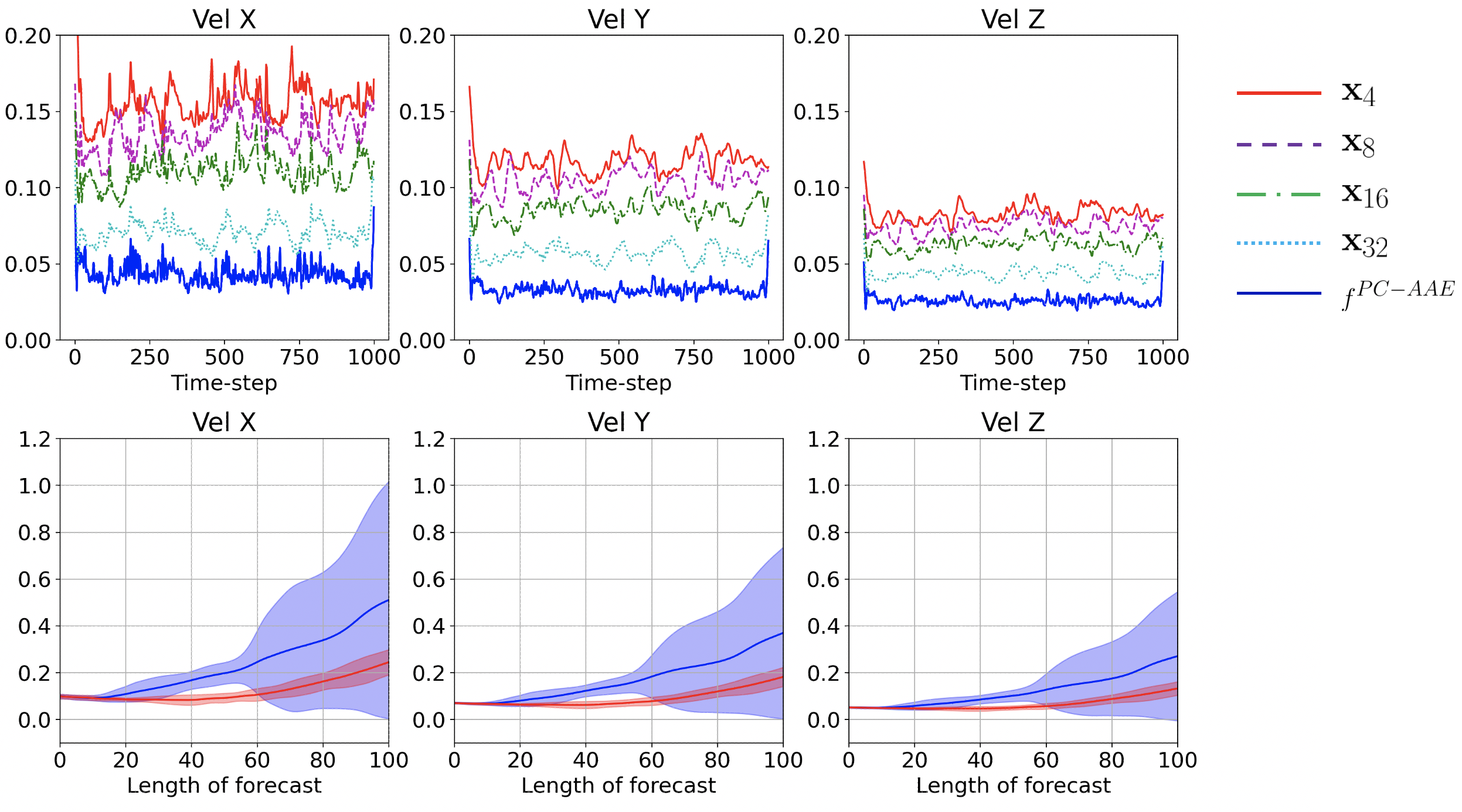}
\caption{Mean absolute error in $m s^{-1}$. Top: $f^{PC-AAE}$ and reconstruction $\mathbf{x}_{\tau}$ with $\tau = \{4, 8, 16, 32\}$ principal components. Bottom: Ensemble of forecast errors with $f^{Classic\,LSTM}$ (blue) and $f^{ALSTM}$ (red) from different starting points from $t=200$ to $t=250$ using 8 dimensions in the latent space. The solid line is the mean and the shaded area is one standard deviation.}
\label{fig:error_aae}
\end{figure*}

Once the latent space $\mathbf{z}$ is obtained, it can be used to train a LSTM \citep{hochreiter1997long} to make predictions. In this paper, an adversarial LSTM (ALSTM) network takes $N$ previous time-levels of $\mathbf{z}_{t_{k-N}},\dots, \mathbf{z}_{t_{k}}$ as input and predicts an approximation of $\mathbf{z}_{t_{k+1}}$ named $\tilde{\mathbf{z}}_{t_{k+1}}$. The functional of ALSTM is defined as:

\begin{equation}
f^{ALSTM}:\mathbf{z}_{t_{k-N}},\dots, \mathbf{z}_{t_{k}} \to  \tilde{\mathbf{z}}_{t_{k+1}}
\end{equation}
Our methodology proposes to add adversarial training to the LSTM network to further improve the forecasts. Similar to the adversarial autoencoder described in \ref{sec:aae}, the adversarial training of the LSTM network $\mathcal{G}$ includes a mirrored LSTM discriminator $\mathcal{D}^{B}$ (see Figure \ref{fig:workflow}). The LSTM network $\mathcal{G}$ is designed to predict $\Delta\tilde{\mathbf{z}}_{t_{k+1}} = \mathcal{G}(\mathbf{z}_{t_{k-N}},\dots, \mathbf{z}_{t_{k}})$, with $\Delta\tilde{\mathbf{z}}_{t_{k}} = \tilde{\mathbf{z}}_{t_{k+1}} - \mathbf{z}_{t_{k}}$. Similar to $\mathcal{D}^{A}$, $\mathcal{D}^{B}$ takes $\Delta\mathbf{z}_{t_{k}} = \mathbf{z}_{t_{k+1}} - \mathbf{z}_{t_{k}}$ as a positive sample, and $\Delta\tilde{\mathbf{z}}_{t_{k}}$ as a negative sample. The training of $\mathcal{D}^{B}$ is based on the binary cross-entropy loss, with Nadam, as follows:

\begin{align}
    \mathcal{L}^{adv}_{\mathcal{D}^{B}}(\Delta\mathbf{z}) &= \mathcal{L}^{bce}(\mathcal{D}^{B}(\Delta\mathbf{z}, 1)) + \mathcal{L}^{bce}(\mathcal{D}^{B}(\Delta\tilde{\mathbf{z}}, 0))\\
    \mathcal{L}^{adv}_{f^{ALSTM}}(\Delta\mathbf{z}) &= \mathcal{L}^{bce}(\mathcal{D}^{B}(\Delta\tilde{\mathbf{z}}, 1)) + \mathcal{L}^{mse}(\Delta\tilde{\mathbf{z}}, \Delta\mathbf{z})
\end{align}
Thus, the prediction of the next time-level in the latent space $\tilde{\mathbf{z}}_{t_{k+1}}$ comes from $\tilde{\mathbf{z}}_{t_{k+1}} = \Delta\tilde{
\mathbf{z}}_{t_{k}} + \mathbf{z}_{t_{k}}$. Once both networks are trained, the prediction of the next time-level in the physical space $\tilde{\mathbf{x}}_{t_{k+1}}$  is given by $\tilde{\mathbf{x}}_{t_{k+1}} = \mathcal{P}(\tilde{\mathbf{z}}_{t_{k+1}}) + \bar{\mathbf{x}}$.

\subsection{Domain}

The computational fluid dynamics (CFD) simulations were carried out using Fluidity \citep{davies2011fluidity}. The 3D case is a realistic case including 14 buildings representing a real urban area located near Elephant and Castle, South London, UK. The 3D case ($720m\times676m\times250m$) is composed of an unstructured mesh including $m = 100,040$ nodes per dimension and $n = 1000$ time-steps 

A log-profile velocity (representing the wind profile of the atmospheric boundary layer) is used in the 3D case. The building facades and bottom surface have no-slip boundary conditions. Whilst, top and sides of the domain have a free-slip boundary condition. A more detailed description can be found in \citet{arcucci2019optimal}.
\section{Results and discussion}

\begin{figure*}[t!]
\centering
\includegraphics[trim = {0cm 2.5cm 0cm 2.5cm},clip,width =0.9\textwidth]{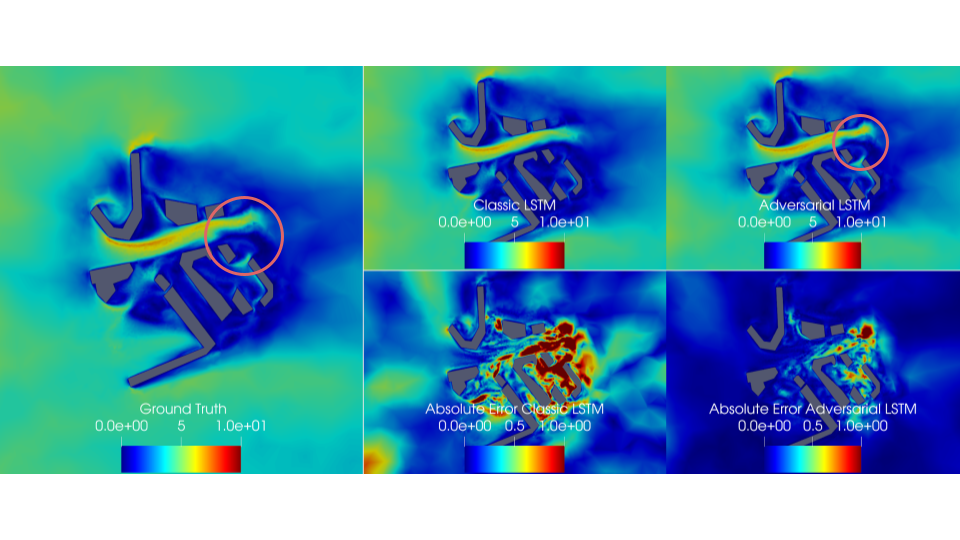}
\caption{Comparison of forecasted velocity in $m s^{-1}$ (magnitude) fields by $f^{Classic\,LSTM}$ and $f^{ALSTM}$ and the absolute error with respect to the ground truth. This is a 80 time-step forecast starting from $t=200$. This is a horizontal slice normal to the Z-axis 10 m above ground.}
\label{fig:vel_fields}
\end{figure*}

A PCA was applied to a 3-dimensional velocity field $(m s^{-1})$. The full-rank PCs were used as input for the AAE and divided in 4 different experiments named $LS_{\tau}$  and compared to the corresponding reconstruction $\mathbf{x}_{\tau}$ with $\tau = \{4, 8, 16, 32\}$ PCs. The results of the mean absolute error using the different dimension reduction approaches are shown in Figure \ref{fig:error_aae}. The AAE outperforms a simple truncation of the PCs. Additionally, the solution is robust and shows similar results regarding how many dimensions are chosen in the latent space of the bottleneck layer of the AAE. Therefore, the following forecast results are based on a dimension reduction using 8 dimensions in the latent space of the AAE, which is a compression of 5 orders of magnitude. The chosen architectures, hyperparameters and training options are shown in Appendix \ref{app:1}.

Figure \ref{fig:error_aae} presents the error from an ensemble of forecasts, of velocities in X, Y, Z, starting from different time-steps. The solid line represents the mean of the error ensemble and the shaded area is the standard deviation. The forecasts are created by using previous time-steps from data and producing a forecast which is subsequently used as an input for the prediction of the next time-step. The forecast experiments are named $F_{\tau}$. After 100 iterations, it is very clear that $f^{ALSTM}$ outperforms a LSTM with the same architecture without adversarial training, named $f^{Classic\,LSTM}$. The forecasts are 4 orders of magnitude faster than the CFD simulation.

Figure \ref{fig:vel_fields} shows the comparison of forecasted magnitude velocity (in $m s^{-1}$) fields of $f^{ALSTM}$, with 8 dimensions in the latent space, and $f^{Classic\,LSTM}$, with 8 PCs, from $t=200$. The snapshots clearly show that after 80 time-levels of forecasting, $f^{Classic\,LSTM}$ diverges quickly from the underlying model state, while $f^{ALSTM}$ preserve more underlying physics. Furthermore, $f^{ALSTM}$ is able to recreate eddies accurately learnt from the underlying physics model (red circle in Figure \ref{fig:vel_fields}). This approach outperforms \citep{casas2020reduced} in terms of data compression and \citep{quilodran2021adversarially} in terms of forecast length.

\section{Conclusions}

This paper presented the advantages of using adversarial training to improve the forecast or a CFD simulation of urban air pollution. The PC-based adversarial autoencoder architecture is robust and its dimension-reduction outperforms a simple truncation of PCs. Furthermore, it is shown that applying adversarial training to a LSTM outperforms a LSTM trained in a classical way. This is very important when accurate near real-time predictions are needed and not enough data is available. It can be observed that adversarially trained LSTM does not diverge greatly from the data it has learned, given the constraint of the discriminator network. The replacement of the CFD solution by these models will speed up the forecast process towards a real-time solution. And, the application of adversarial training could potentially produce more physically realistic flows. Furthermore, this framework is data-agnostic and could be applied to different CFD models where enough data is available.

\clearpage
\section*{Acknowledgements}

This work is supported by the EPSRC Grand Challenge grant ‘Managing Air for Green Inner Cities (MAGIC) EP/N010221/1, EP/T000414/1 PREdictive Modelling with QuantIfication of UncERtainty for MultiphasE Systems (PREMIERE), the EP/T003189/1 Health assessment across biological length scales for personal pollution exposure and its mitigation (INHALE), the RELIANT grant (EP/V036777/1) and the Leonardo Centre for Sustainable Business at Imperial College London.

\bibliography{iclr2021_conference}
\bibliographystyle{iclr2021_conference}
\clearpage
\appendix
\section{Appendix}
\label{app:1}
\begin{table*}[ht]
\centering
  \caption{Architectures of the different networks used in this study for the adversarial autoencoders and the adversarial LSTMs. The encoder $\mathcal{Q}$, the decoder $\mathcal{P}$ and the discriminator $\mathcal{D}^{A}$ are fully-connected layers. In the forecasts experiments Generator $\mathcal{G}$ and Discriminator $\mathcal{D}^{B}$ are LSTM networks.}
  \label{table:hyperparameters}
  \begin{tabular}{|c|c|c|c|c|c|c|}
    \hline
    Experiments AE & Enc $\mathcal{Q}$ & Dec $\mathcal{P}$ & Disc $\mathcal{D^{A}}$ & Experiments ALSTM & Gen $\mathcal{G}$ & Disc $\mathcal{D}^{B}$\\
    \hline
    $LS_{32}$ & 1000 & 32 & 32 & $F_{32}$ & 32 & 32\\
    & 64 & 64 & 1 & & 64 & 64\\
    & 32 & 1000 & & & 32 & 1\\
    \hline
    $LS_{16}$ & 1000 & 16 & 16 & $F_{16}$ & 16 & 16\\
    & 64 & 32 & 1 & & 64 & 64\\
    & 32 & 64 & & & 16 & 1 \\
    & 16 & 1000 & & & & \\
    \hline
    $LS_{8}$ & 1000 & 8 & 8 & $F_{8}$ & 8 & 8\\
    & 64 & 16 & 1 & & 64 & 64\\
    & 32 & 32 & & & 8 & 1\\
    & 16 & 64 & & & &\\
    & 8 & 1000 & & & &\\
    \hline
    $LS_{4}$ & 1000 & 4 & 4 & $F_{4}$ & 4 & 4\\
    & 64 & 8 & 1 & & 64 & 64\\
    & 32 & 16 & & & 4 & 1\\
    & 16 & 32 & & & &\\
    & 8 & 64 & & & &\\
    & 4 & 1000 & & & &\\
    \hline
    \multicolumn{4}{|c|}{Batch Normalisation in-between layers} & \multicolumn{3}{|c|}{Batch Normalisation before final layer}\\
    \multicolumn{4}{|c|}{Leaky ReLU with a slope of 0.3} & \multicolumn{3}{|c|}{Dropout 0.5}\\
    \multicolumn{4}{|c|}{} & \multicolumn{3}{|c|}{Time-lag $= 5$}\\
    \hline
    \multicolumn{7}{|c|}{Optimizer: Adam with Nesterov momentum with Learning rate $=10^{-3}, \beta_1=0.9, \beta_2=0.999$}\\
    \hline
    \end{tabular}
\end{table*}

\end{document}